\documentclass[11pt,a4paper]{article}
\usepackage[hyperref]{acl2019}
\usepackage{times}
\usepackage{latexsym}
\usepackage{bbm}
\usepackage{algorithm}
\usepackage{amsmath,amsthm,amssymb,bbm}
\usepackage{mathtools}
\usepackage{cases}
\usepackage{comment}
\usepackage{url}
\usepackage{booktabs}
\usepackage{multirow}
\usepackage{color}
\usepackage{bbm}
\usepackage{arydshln}
\usepackage{caption}
\usepackage{subcaption}

\usepackage{url}

\usepackage{algpseudocode,algorithm,algorithmicx}

\algrenewcommand\algorithmicrequire{\textbf{Precondition:}}
\algrenewcommand\algorithmicensure{\textbf{Postcondition:}}

\newcommand{\unk}{\colorbox{gray!20}{\texttt{UNK}}}
\newcommand{\unkspace}{\colorbox{gray!20}{\texttt{UNK}} }

\title{
Combating Adversarial Misspellings
with Robust Word Recognition
}

\author{Danish Pruthi \qquad Bhuwan Dhingra \qquad Zachary C. Lipton \\
    Carnegie Mellon University \\ Pittsburgh, USA \\
    \texttt{\{ddanish, bdhingra\}@cs.cmu.edu}, \texttt{zlipton@cmu.edu}
}

\aclfinalcopy
\date{}

\begin{document}
\maketitle

\begin{abstract}
To combat adversarial spelling mistakes,
we propose placing a word recognition model
in front of the downstream classifier.
Our word recognition models build upon
the RNN semi-character architecture,
introducing several new \emph{backoff} strategies
for handling rare and unseen words.
Trained to recognize words corrupted by random adds,
drops, swaps, and keyboard mistakes,
our method achieves $32\%$ relative (and $3.3\%$ absolute)
error reduction over the vanilla semi-character model.
Notably, our pipeline confers robustness on the downstream classifier,
outperforming both adversarial training and off-the-shelf spell checkers.
Against a BERT model fine-tuned for sentiment analysis,
a single adversarially-chosen character attack
lowers accuracy from $90.3\%$ to $45.8\%$.
Our defense restores accuracy to $75\%$\footnote{All code for our defenses, attacks, and baselines is available at \url{https://github.com/danishpruthi/adversarial-misspellings}}.
Surprisingly, better word recognition
does not always entail greater robustness.
Our analysis reveals that robustness also depends
upon a quantity that we denote the \emph{sensitivity}.
\end{abstract}

\section{Introduction}
\label{sec:intro}
Despite the rapid progress of deep learning techniques on diverse supervised learning tasks,
these models remain brittle to subtle shifts in the data distribution.
Even when
the permissible changes
are confined to barely-perceptible perturbations,
training robust models remains an open challenge.
Following the discovery that imperceptible attacks
could cause image recognition models to misclassify examples \citep{szegedy2013intriguing},
a veritable sub-field has emerged in which authors
iteratively propose attacks and countermeasures.

For all the interest in adversarial computer vision,
these attacks are rarely encountered outside of academic research.
However, adversarial misspellings constitute a \emph{longstanding real-world problem}.
Spammers continually bombard email servers,
subtly misspelling words in efforts to evade spam detection
while preserving the emails' intended meaning \citep{lee2005spam,fumera2006spam}.
As another example, programmatic censorship on the Internet
has spurred communities to adopt similar methods
to communicate surreptitiously \citep{bitso2013trends}.

\begin{table}
\small
\centering
\begin{tabular}{@{}ccc@{}}
\toprule
\textbf{Alteration}                    & \textbf{Movie Review}                                                                                            & \textbf{Label} \\ \midrule
Original                      & \begin{tabular}[c]{@{}c@{}}A triumph, relentless and beautiful\\  in its downbeat darkness\end{tabular} & \textbf{+}     \\ \midrule
Swap                     & \begin{tabular}[c]{@{}c@{}}A triumph, relentless and \textcolor{red}{beuatiful}\\  in its downbeat darkness\end{tabular} & \textbf{--}     \\
Drop                     & \begin{tabular}[c]{@{}c@{}}A triumph, relentless and beautiful\\  in its \textcolor{red}{dwnbeat} darkness\end{tabular} & \textbf{--}     \\ \midrule
+ Defense                     & \begin{tabular}[c]{@{}c@{}}A triumph, relentless and \textcolor{blue}{beautiful}\\ in its downbeat darkness\end{tabular}  & \textbf{+}     \\
+ Defense & \begin{tabular}[c]{@{}c@{}}A triumph, relentless and beautiful\\ in its \textcolor{blue}{downbeat} darkness\end{tabular}  & \textbf{+}     \\ \bottomrule
\end{tabular}
\caption{Adversarial spelling mistakes inducing sentiment misclassification and word-recognition defenses.}
\end{table}

In this paper, we focus on adversarially-chosen spelling mistakes
in the context of text classification, addressing the following attack types:
dropping, adding, and swapping internal characters within words.
These perturbations are inspired by psycholinguistic studies \cite{rawlinson1976significance, davis2013cu}
which demonstrated that humans can comprehend text altered by jumbling internal characters,
provided that  the first and last characters of each word remain unperturbed.

First, in experiments addressing both BiLSTM and fine-tuned BERT models,
comprising four different input formats:
word-only, char-only, word+char, and word-piece \citep{wu2016google},
we demonstrate that an adversary can degrade a classifier's performance
to that achieved by random guessing.
\emph{This requires altering just two characters per sentence}.
Such modifications might flip words either to a different word in the vocabulary
or, more often, to the out-of-vocabulary token \unk.
Consequently, adversarial edits can degrade a word-level model
by transforming the informative words to \unk.
Intuitively, one might suspect that word-piece and character-level models
would be less susceptible to spelling attacks
as they can make use of the residual word context.
However, our experiments demonstrate that character and word-piece models
are in fact \emph{more vulnerable.}
We show that this is due to the adversary's effective capacity
for finer grained manipulations on these models.
While against a word-level model, the adversary is mostly limited to \unk-ing words, against a word-piece or character-level model,
each character-level add, drop, or swap produces a distinct input,
providing the adversary with a greater set of options.

Second, we evaluate first-line techniques including
data augmentation and adversarial training,
demonstrating that they offer only marginal benefits here, e.g., a BERT model achieving $90.3$ accuracy on a sentiment classification task,
is degraded to $64.1$ by an adversarially-chosen
$1$-character swap in the sentence,
which can only be restored to $69.2$ by adversarial training.

Third (our primary contribution),
we propose a task-agnostic defense, attaching
a word recognition model that predicts each word in a sentence given
a full sequence of (possibly misspelled) inputs.
The word recognition model's outputs form
the input to a downstream classification model.
Our word recognition models build upon the RNN-based semi-character
word recognition model due to \citet{sakaguchi2017robsut}.
While our word recognizers are trained on domain-specific text
from the task at hand, they often predict \unkspace at test time,
owing to the small domain-specific vocabulary.
To handle unobserved and rare words,
we propose several \emph{backoff} strategies
including falling back on a generic word recognizer trained on a larger corpus.
Incorporating our defenses, BERT models subject to 1-character attacks are restored to $88.3$, $81.1$, $78.0$ accuracy
for swap, drop, add attacks respectively,
as compared to $69.2$, $63.6$, and $50.0$ for adversarial training

Fourth, we offer a detailed qualitative analysis,
demonstrating that a low word error rate alone is insufficient for a word recognizer to confer robustness on the downstream task.
Additionally, we find that it is important that the recognition model
supply few degrees of freedom to an attacker.
We provide a metric to quantify this notion of \emph{sensitivity}
in word recognition models
and study its relation to robustness empirically.
Models with low sensitivity \emph{and} word error rate are most robust.

\section{Related Work}
\label{sec:related}

Several papers address adversarial attacks on NLP systems.
Changes to text, whether word- or character-level, are all perceptible,
raising some questions about what should rightly be considered an adversarial example~\citep{ebrahimi2017hotflip, belinkov2017synthetic}.
\citet{jia2017adversarial} address the reading comprehension task,
showing that by appending \emph{distractor sentences}
to the end of stories from the SQuAD dataset \cite{rajpurkar2016squad},
they could cause models to output incorrect answers.
Inspired by this work, \citet{glockner2018breaking}
demonstrate an attack that breaks entailment systems
by replacing a single word with either a synonym or its hypernym.
Recently, \citet{zhao2018generating}
investigated the problem of producing natural-seeming adversarial examples,
noting that adversarial examples in NLP
are often ungrammatical \citep{li2016understanding}.

In related work on character-level attacks, \citet{ebrahimi2017hotflip, ebrahimi2018adversarial}
explored gradient-based methods to generate string edits
to fool classification and translation systems, respectively.
While their focus is on efficient methods for generating
adversaries, ours is on improving the worst case
adversarial performance.
Similarly, \citet{belinkov2017synthetic} studied
how synthetic and natural noise affects character-level machine translation.
They considered structure invariant representations and
adversarial training as defenses against such noise.
Here, we show that an auxiliary word recognition model,
which can be trained on unlabeled data,
provides a strong defense.

Spelling correction~\citep{kukich1992techniques}
is often viewed as a sub-task of grammatical error correction \citep{ng2014conll,schmaltz-EtAl:2016:BEA11}.
Classic methods rely on a source language model
and a noisy channel model
to find the most likely correction for a given word \citep{MAYS1991517,Brill:2000:IEM:1075218.1075255}.
Recently, neural techniques have been applied to the task \citep{sakaguchi2017robsut,li2018spelling},
which model the context and orthography of the input together.
Our work extends the ScRNN model of \citet{sakaguchi2017robsut}.

\section{Robust Word Recognition}
\label{sec:methodology}
To tackle character-level adversarial attacks,
we introduce a simple two-stage solution,
placing a word recognition model ($W$)
before the downstream classifier ($C$).
Under this scheme, all inputs are classified
by the composed model $C \circ W$.
This modular approach, with $W$ and $C$ trained separately, offers several benefits:
(i) we can deploy the same word recognition model for multiple downstream classification tasks/models;
and (ii) we can train the word recognition model with larger unlabeled corpora.

Against adversarial mistakes,
two important factors
govern the robustness of this combined model:
$W$'s \emph{accuracy} in recognizing misspelled words
and $W$'s \emph{sensitivity} to adversarial perturbations on the same input.
We discuss these aspects in detail below.

\begin{figure*}
    \centering
    \includegraphics[width=0.7\textwidth]{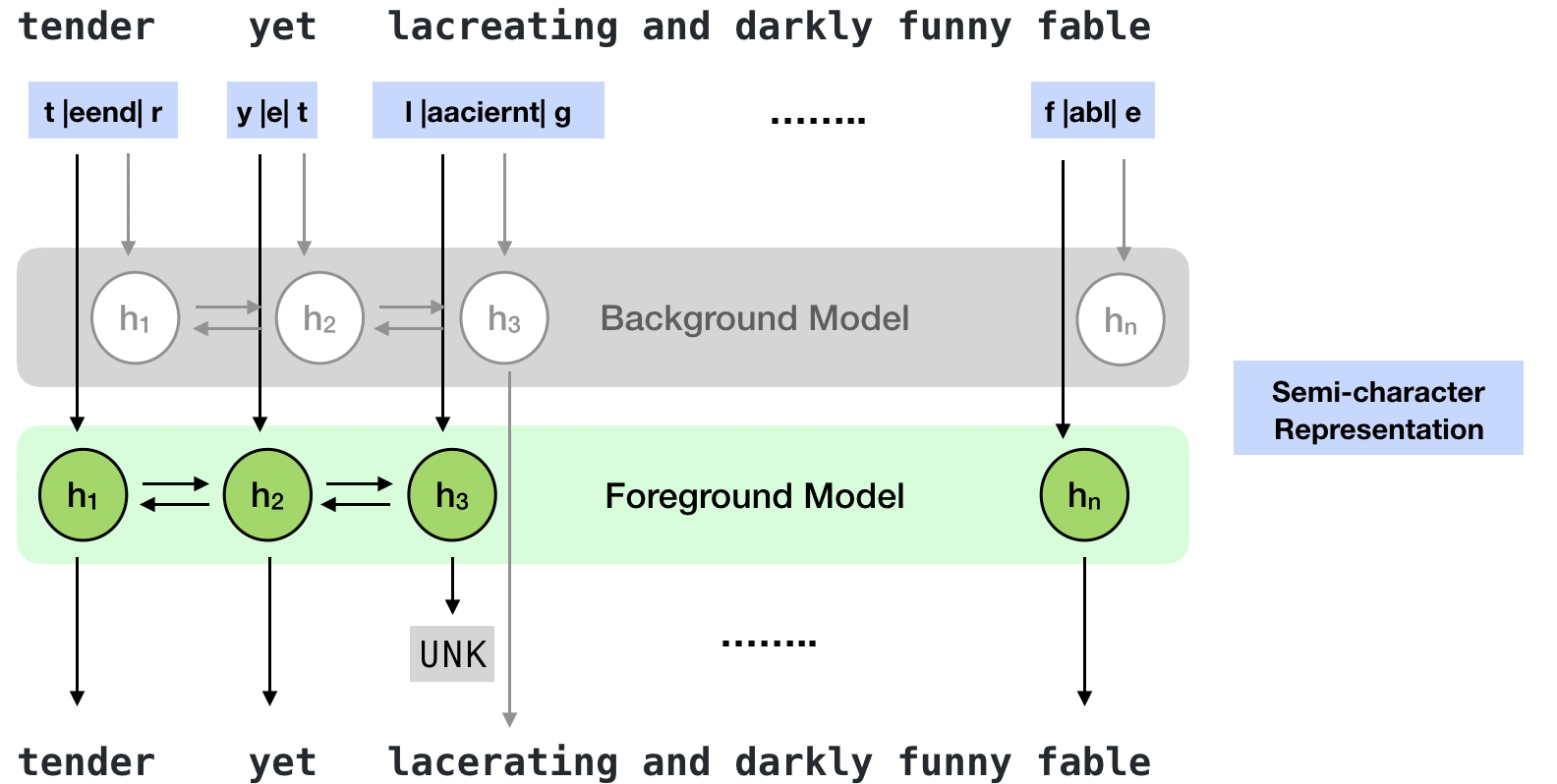}
    \caption{A schematic sketch of our proposed
    word recognition system,
    consisting of a \emph{foreground} and a \emph{background} model.
    We train the foreground model on the smaller,  domain-specific dataset,
    and the background model on a larger dataset (e.g., the IMDB movie corpus).
    We train both models to reconstruct the correct word from the orthography and context of the individual words, using synthetically corrupted inputs during training.
    Subsequently, we invoke the background model
    whenever the foreground model predicts \unk.}
    \label{fig:overall}
\end{figure*}

\subsection{ScRNN with Backoff}
We now describe semi-character RNNs for word recognition,
explain their limitations, and suggest techniques to improve them.

\paragraph{ScRNN Model}
Inspired by the psycholinguistic studies~\citep{davis2013cu, rawlinson1976significance},
\citet{sakaguchi2017robsut} proposed a semi-character based RNN (ScRNN)
that processes a sentence of words with misspelled characters,
predicting the correct words at each step.
Let $s = \{w_1, w_2, \dots, w_n\}$ denote the input sentence,
a sequence of constituent words $w_i$.
Each input word ($w_i$) is represented by concatenating
(i) a one hot vector of the first character ($\mathbf{w_{i1}}$);
(ii) a one hot representation of the last character
($\mathbf{w_{il}}$, where $l$ is the length of word $w_i$);
and (iii) a bag of characters representation of the internal characters ($\sum_{j=2}^{l-1}\mathbf{w_{ij}})$.
ScRNN treats the first and the last characters individually,
and is agnostic to the ordering of the internal characters.
Each word, represented accordingly,
is then fed into a BiLSTM cell.
At each sequence step, the training target is the correct corresponding word
(output dimension equal to vocabulary size),
and the model is optimized with cross-entropy loss.

\paragraph{Backoff Variations}
While~\citet{sakaguchi2017robsut} demonstrate strong word recognition performance,
a drawback of their evaluation setup
is that they only attack and evaluate
on the subset of words that are a part of their training vocabulary.
In such a setting, the word recognition performance
is unreasonably dependent on the chosen vocabulary size.
In principle,
one can  design models to predict
(correctly)
only a few chosen words,
and ignore the remaining majority and still reach $100$\% accuracy.
\emph{For the adversarial setting,
rare and unseen words in the wild are particularly critical,
as they provide opportunities for the attackers.}
A reliable word-recognizer should handle these cases gracefully.
Below, we explore different ways to
\emph{back off} when the ScRNN predicts \unkspace (a frequent outcome for rare and unseen words):
\begin{itemize}
    \item  \textbf{Pass-through}: word-recognizer passes on the (possibly misspelled) word as is.
    \item \textbf{Backoff to neutral word}:
    Alternatively, noting that passing $\unk$-predicted words through unchanged exposes the downstream model to potentially corrupted text,
    we consider backing off to a neutral word like `a',
    which has a similar distribution across classes.
    \item \textbf{Backoff to background model}:
    We also consider falling back upon a more generic word recognition model
    trained upon a larger, less-specialized corpus
    whenever the foreground word recognition model predicts \unk \footnote{Potentially the background model could be trained with full vocabulary so that it never predicts \unk}. Figure~\ref{fig:overall} depicts this scenario pictorially.
\end{itemize}

\noindent Empirically, we find that the background model (by itself) is less accurate, because of the large number of words it is trained to predict.
Thus, it is best to train a precise foreground model
on an in-domain corpus and focus on frequent words,
and then to resort to a general-purpose background model for
rare and unobserved words.
Next, we delineate our second consideration for building robust word-recognizers.

\subsection{Model Sensitivity}
\label{sec:sensitivity}

In computer vision, an important factor determining the success of an adversary is the norm constraint on the perturbations allowed to an image ($|| \bf x - \bf x'||_{\infty} < \epsilon$). Higher values of $\epsilon$ lead to a higher chance of mis-classification for at least one $\bf x'$. Defense methods such as quantization \citep{xu2017feature} and thermometer encoding \citep{buckman2018thermometer} try to reduce the space of perturbations available to the adversary by making the model invariant to small changes in the input.

In NLP, we often get such invariance for free,
e.g., for a word-level model, most of the perturbations
produced by our character-level adversary
lead to an \unkspace at its input.
If the model is robust to the presence of these \unkspace tokens,
there is little room for an adversary to manipulate it.
Character-level models, on the other hand,
despite their superior performance in many tasks,
do not enjoy such invariance.
This characteristic invariance could be exploited by an attacker.
Thus, to limit the number of different inputs to the classifier,
we wish to reduce the number of distinct word recognition outputs
that an attacker can induce, not just the number of words on which the model is ``fooled''.
We denote this property of a model as its \emph{sensitivity}.

We can quantify this notion for
a word recognition system $W$ as
the expected number of unique outputs
it assigns to a set of adversarial perturbations.
Given a sentence $s$ from the set of sentences $\mathcal{S}$,
let $A(s) = {s_1}' , {s_2}', \dots, {s_n}'$
denote the set of $n$ perturbations to it under attack type $A$,
and let $V$ be the function that maps
strings to an input representation
for the downstream classifier.
For a word level model, $V$ would transform
sentences to a sequence of word ids,
mapping OOV words to the same \unkspace ID.
Whereas, for a char (or word+char, word-piece) model, $V$ would map inputs to a
sequence of character IDs.
Formally, sensitivity is defined as

\begin{equation}
     S_{W,V}^A=\mathbb{E}_{s}\left[\frac{\#_{u}(V \circ W({s_1}'), \dots, V \circ W({s_n}'))}{n}\right] ,
     \label{eq:sensitivity}
\end{equation}
where $V \circ W (s_i)$ returns the input representation (of the downstream classifier) for the output string produced by the word-recognizer $W$ using $s_i$
and $\#_{u}(\cdot)$ counts the number of unique arguments.

Intuitively, we expect a high value of $S_{W, V}^A$ to lead to a lower robustness of the downstream classifier, since the adversary has more degrees of freedom to attack the classifier. Thus, when using word recognition as a defense,
it is prudent to design a low sensitivity system with a low error rate.
However, as we will demonstrate,
there is often a trade-off between sensitivity
and error rate.

\subsection{Synthesizing Adversarial Attacks}

Suppose we are given a classifier $C: \mathcal{S} \to \mathcal{Y}$ which maps natural language sentences $s \in \mathcal{S}$ to a label from a predefined set $y \in \mathcal{Y}$. An adversary for this classifier is a function $A$ which maps a sentence $s$ to its perturbed versions $\{s'_1, s'_2, \ldots, s'_{n}\}$ such that each $s'_i$ is close to $s$ under some notion of distance between sentences.
We define the robustness of classifier $C$ to the adversary $A$ as:
\begin{equation}
    R_{C,A} = \mathbb{E}_s \left[\min_{s' \in A(s)} \mathbbm{1}[C(s') = y]\right],
    \label{eq:robustness}
\end{equation}
where $y$ represents the ground truth label for $s$.
In practice, a real-world adversary may only be able to query the classifier a few times, hence $R_{C,A}$ represents the \textit{worst-case} adversarial performance of $C$.
Methods for generating adversarial examples,
such as HotFlip \citep{ebrahimi2017hotflip},
focus on efficient algorithms for searching the $\min$ above.
Improving $R_{C,A}$ would imply better robustness against all these methods.

\paragraph{Allowed Perturbations $(A(s))$} We explore adversaries which perturb sentences with four types of character-level edits:
\noindent (1) Swap: swapping two adjacent internal characters of a word.
(2) Drop: removing an internal character of a word.
(3) Keyboard: substituting an internal character with adjacent characters of QWERTY keyboard
(4) Add: inserting a new character internally in a word.
In line with the psycholinguistic studies~\cite{davis2013cu, rawlinson1976significance}, to ensure that the perturbations do not affect human ability to comprehend the sentence, we only allow the adversary to edit the internal characters of a word, and not edit stopwords or words shorter than $4$ characters.

\paragraph{Attack Strategy}
For \textit{1-character} attacks, we
try all possible perturbations listed above
until we find an adversary that flips the model prediction.
For \textit{2-character} attacks,
we greedily fix the edit which had the least confidence
among 1-character attacks,
and then
try all the allowed perturbations on the remaining words.
Higher order attacks can be performed in a similar manner.
The greedy strategy reduces the computation required to obtain higher order attacks\footnote{Its complexity is $O(l)$, instead of $O(l^m)$ where $l$ is the sentence length and $m$ is the order.},
but also means that the robustness score is an upper bound on the true robustness of the classifier.

\section{Experiments and Results}
\label{sec:experiments}

In this section, we first discuss our experiments on the word recognition systems.

\subsection{Word Error Correction}

\textbf{Data}: We evaluate the spell correctors from \S\ref{sec:methodology}
on movie reviews from the Stanford Sentiment Treebank (SST) \citep{socher2013recursive}.
The SST dataset consists of $8544$ movie reviews,
with a vocabulary of over 16K words.
As a background corpus, we use the IMDB movie reviews \citep{maas-EtAl:2011:ACL-HLT2011},
which contain $54$K movie reviews,
and a vocabulary of over 78K words.
The two datasets do not share any reviews in common.
The spell-correction models are evaluated
on their ability to correct misspellings.
The test setting consists of reviews where each word
(with length $\ge 4$, barring stopwords)
is attacked by one of the attack types
(from swap, add, drop and keyboard attacks).
In the \emph{all} attack setting,
we mix all attacks by randomly choosing one for each word.
This most closely resembles a real world attack setting.

\paragraph{Experimental Setup}
In addition to our word recognition models,
we also compare to After The Deadline (ATD),
an open-source spell corrector\footnote{\url{https://www.afterthedeadline.com/}}.
We found ATD to be the best freely-available corrector\footnote{We compared ATD with Hunspell (\url{http://hunspell.github.io/}), which is used in Linux applications.
ATD was significantly more robust owing to taking context into account while correcting.}.
We refer the reader to \citet{sakaguchi2017robsut}
for comparisons of ScRNN to other anonymized commercial spell checkers.

For the ScRNN model, we use a single-layer Bi-LSTM
with a hidden dimension size of $50$.
The input representation consists of $198$ dimensions,
which is thrice the number of unique characters ($66$) in the vocabulary.
We cap the vocabulary size to $10$K words,
whereas we use the entire vocabulary of $78470$ words
when we backoff to the background model.
For training these networks,
we corrupt the movie reviews according to all attack types,
i.e., applying one of the $4$ attack types to each word,
and trying to reconstruct the original words via cross entropy loss.

\begin{table}[hbt!]
\small
    \centering
    \begin{tabular}{lccccc}
         \multicolumn{6}{c}{\textbf{Word Recognition}}\\
         \toprule
        Spell-Corrector & \bf{Swap} & \bf{Drop} & \bf{Add} & \bf{Key} & \bf{All} \\
        \midrule
        ATD & 7.2 & 12.6 & 13.3 & 6.9 & 11.2 \\ \midrule
        ScRNN ($78$K) & 6.3 & 10.2 & 8.7 & 9.8 & 8.7 \\\midrule
         \multicolumn{6}{c}{ScRNN ($10$K) w/ Backoff Variants}\\ \midrule
        Pass-Through & 8.5 &	10.5 &	10.7 &	11.2 &	10.2 \\
        Neutral & 8.7 & 10.9 & 10.8 & 11.4 & 10.6\\
        Background & \textbf{5.4} &\textbf{8.1} & \textbf{6.4} & \textbf{7.6} & \textbf{6.9} \\
    \hline
    \end{tabular}
    \caption{
    Word Error Rates (WER) of ScRNN with each backoff strategy, plus ATD
    and an ScRNN trained only on the background corpus ($78$K vocabulary)
    The error rates include 5.25\% OOV words.
    }
    \label{tab:wer-results}
\end{table}

 \paragraph{Results}
We calculate the word error rates (WER) of each of the models for different attacks
and present our findings in  Table~\ref{tab:wer-results}.
Note that ATD incorrectly predicts $11.2$ words for every $100$ words (in the `all' setting),
whereas, all of the backoff variations of the ScRNN reconstruct better.
The most accurate variant involves backing off to
the background model, resulting in a low error rate of $6.9\%$,
leading to the best performance on word recognition.
This is a $32\%$ relative 
error reduction
compared to the vanilla ScRNN model
with a pass-through backoff strategy.
We can attribute the improved performance to
the fact that there are $5.25\%$ words in the test corpus that are unseen in the training corpus,
and are thus only recoverable by backing off to a larger corpus.
Notably, only training on the larger background corpus does worse,
at $8.7\%$, since the distribution
of word frequencies is different in the background corpus
compared to the foreground corpus.

\subsection{Robustness to adversarial attacks}

\begin{table*}[hbt!]
\small
    \centering
    \begin{tabular}{lcccccc}
         \multicolumn{7}{c}{\textbf{Sentiment Analysis} ($1$-char attack/$2$-char attack)}\\
         \toprule
        \textbf{Model} & \bf{No attack} & \bf{Swap} & \bf{Drop} & \bf{Add} & \bf{Key} & \bf{All} \\
        \midrule
        \multicolumn{7}{c}{Word-Level Models}\\
        \midrule
        BiLSTM & 79.2 &  (64.3/53.6) &	(63.7/52.7)	& (60.0/43.2)	& (60.2/42.4)		& (58.6/40.2) \vspace{2mm} \\
        BiLSTM + ATD & 79.3	 & (76.2/75.3) & (66.5/59.9) & (55.6/47.5) & (62.6/57.6) & (55.8/37.0) \\

        BiLSTM + Pass-through  & 79.3 & (78.6/78.5)	& (69.1/65.3) & (65.0/59.2) & (\textbf{69.6/65.6}) & (63.2/52.4) \\
        BiLSTM + Background   & 78.8	& (78.9/78.4)		& (69.6/66.8) & (62.6/56.4)	 & (68.2/62.2) & (59.6/49.0)\\
        BiLSTM + Neutral  & 80.1	& \textbf{(80.1/79.9)} &		\textbf{(72.4/70.2)}	& \textbf{(67.2/61.2)} & (69.0/64.6)	& \textbf{(63.2/54.0)} \\
        \midrule


        \multicolumn{7}{c}{Char-Level Models}\\
        \midrule
        BiLSTM & 70.3 & (53.6/42.9) &	(48.8/37.1) & (33.8/14.8) &	(40.8/22.0) & (32.6/14.0) \vspace{2mm} \\
        BiLSTM + ATD & 71.0 & (66.6/65.2) & (58.0/53.0) & (54.6/44.4) & (\textbf{61.6}/57.5)	&	(46.5/35.4)\\
        BiLSTM + Pass-through & 70.3 & (65.8/62.9) & (58.3/54.2) &  (54.0/44.2) & (58.8/52.4) & (51.6/39.8) \\
        BiLSTM + Background  & 70.1 & (70.3/69.8)  & (60.4/57.7) & (57.4/52.6) & (58.8/54.2) & (53.6/47.2) \\
        BiLSTM + Neutral & 70.7 & \textbf{(70.7/70.7)}  & \textbf{(62.1/60.5)} & \textbf{(57.8/53.6)} & (61.4/\textbf{58.0}) & \textbf{(55.2/48.4)}\\

        \midrule
        \multicolumn{7}{c}{Word+Char Models}\\
        \midrule
        BiLSTM & 80.5 & (63.9/52.3) & (62.8/50.8) & (57.8/39.8) & (58.4/40.8) & (56.6/35.6)\vspace{2mm}\\

        BiLSTM + ATD & 80.8 & (78.0/77.3) & (67.7/60.9) & (55.6/50.5) & \textbf{(68.7/64.6)}  & (48.5/37.4) \\
        BiLSTM + Pass-through & 80.1 & (79.0/78.7) & (69.5/65.7) & (64.0/59.0)	& (66.0/62.0)	& \textbf{(61.5/56.5)}\\
        BiLSTM + Background & 79.5 & (79.6/79.0)  & (69.7/66.7) & (62.0/57.0)	& (65.0/56.5) & (59.4/49.8)\\
        BiLSTM + Neutral & 79.5 & \textbf{(79.5/79.4)}  & \textbf{(71.2/68.8)} & \textbf{(65.0/59.0)} & (65.5/61.5) &		(\textbf{61.5}/55.5)  \\

        \midrule
        \multicolumn{7}{c}{Word-piece Models}\\
        \midrule
        BERT & 90.3	& (64.1/47.4) & (59.2/39.9)	& (46.2/26.4) & (54.3/34.9)	& (45.8/24.6) \vspace{2mm} \\
        BERT + DA & 90.2 & (68.3/50.6)	& (62.7/39.9)	& (43.6/17.0)	&	(57.7/32.4) & (41.0/15.8) \\
        BERT + Adv & 89.6& (69.2/52.9) & (63.6/40.5) & (50.0/22.0)		& (60.1/36.6) & (47.0/20.2) \vspace{2mm} \\

        BERT + ATD & 89.0 & (84.5/84.5) & (73.0/64.0) & (77.0/69.5)	& (\textbf{80.0}/75.0) & (67.0/55.0) \\
        BERT + Pass-through & 89.8	& (85.5/83.9) & (78.9/75.0) & (70.4/64.4)	& (75.3/70.3) &(68.0/58.5)\\
        BERT + Background & 89.3 & \textbf{(89.1/89.1)} & (79.3/76.5) & (76.5/71.0) & (77.5/74.4) & (73.0/67.5) \\
        BERT + Neutral & 88.3 & (88.3/88.3) & \textbf{(81.1/79.5)} & \textbf{(78.0/74.0)} & (78.8/\textbf{76.8}) & \textbf{(75.0/68.0)}  \\
    \bottomrule
    \end{tabular}
    \caption{
    Accuracy of various classification models,
    with and without defenses, under adversarial attacks.
    Even $1$-character attacks significantly degrade
    classifier performance.
    Our defenses confer robustness, recovering over 76\% of the original accuracy, under the `all' setting for all four model classes.
    }

    \label{tab:sent-results}
\end{table*}

We use sentiment analysis and paraphrase detection as downstream tasks,
as for these two tasks, $1$-$2$ character edits do not change the output labels.

\paragraph{Experimental Setup}
For sentiment classification,
we systematically study the effect of
character-level adversarial
attacks
on two architectures and four different input formats.
The first architecture
encodes the input sentence into a sequence of
embeddings,
which are then sequentially processed by a BiLSTM.
The first and last states of the BiLSTM are then used
by the softmax layer to predict the sentiment of the input.
We consider three input formats for this architecture:
(1) Word-only: where the input words are encoded using a lookup
table;
(2) Char-only: where the input words are encoded using a
separate single-layered BiLSTM over their characters;
and (3) Word$+$Char: where the input words are encoded using
a concatenation of (1) and (2)
\footnote{Implementation details: The embedding dimension size for the word, char and word+char models
are $64$, $32$ and $64+32$ respectively, with $64$, $64$ and $128$ set as the hidden dimension sizes for the three models.}.

The second architecture uses the fine-tuned BERT
model \citep{devlin2018bert},
with an input format of word-piece tokenization.
This model has recently set a new state-of-the-art on
several NLP benchmarks, including the sentiment analysis
task we consider here.
All models are trained and evaluated
on the binary version of the sentence-level
Stanford Sentiment Treebank~\cite{socher2013recursive} dataset
with only positive and negative reviews.

We also consider the task of paraphrase detection.
Here too, we make use of the fine-tuned BERT~\cite{devlin2018bert},
which is trained and evaluated on the Microsoft Research Paraphrase Corpus (MRPC)
\citep{dolan2005automatically}.

\paragraph{Baseline defense strategies}
Two common methods for dealing with adversarial examples include:
(1) data augmentation (\textbf{DA}) \citep{krizhevsky2012imagenet};
and (2) adversarial training (\textbf{Adv}) \citep{goodfellow2014explaining}.
In \textbf{DA}, the trained model is fine-tuned
after augmenting the training set
with an equal number of examples
randomly attacked with a 1-character edit.
In \textbf{Adv}, the trained model is fine-tuned
with additional adversarial examples (selected at random)
that produce incorrect predictions from the current-state classifier.
The process is repeated iteratively,
generating and adding newer adversarial examples from the updated classifier model,
until the adversarial accuracy on dev set stops improving.

\paragraph{Results}
In Table~\ref{tab:sent-results}, we examine the robustness
of the sentiment models under each attack and defense method.
In the absence of any attack or defense, BERT (a word-piece model) performs the best ($90.3\%$\footnote{The reported accuracy on SST-B by BERT in Glue Benchmarks is slightly higher as it is trained and evaluated on \textit{phrase-level} sentiment prediction task which has more training examples compared to the \textit{sentence-level} task we consider. We use the official source code at  \url{https://github.com/google-research/bert}})
followed by word+char models ($80.5\%$), word-only models ($79.2\%$) and then char-only models ($70.3\%$).
However,
even single-character attacks (chosen adversarially) can be catastrophic,
resulting in a significantly degraded performance
of $46\%$, $57\%$, $59\%$ and $33\%$, respectively under the `all' setting.

Intuitively, one might suppose
that word-piece and character-level models
would be more robust to such attacks
given they can make use of the remaining context.
However, we find that they are the more susceptible.
To see why, note that the word `beautiful' can only be altered
in a few ways for word-only models,
either leading to an \unkspace or an existing vocabulary word,
whereas, word-piece and character-only models treat each unique character combination differently.
This provides more variations that an attacker can exploit.
Following similar reasoning, \emph{add} and \emph{key} attacks
pose a greater threat than \emph{swap} and \emph{drop} attacks.
The robustness of different models can be ordered
as word-only $>$ word+char $>$ char-only $\sim$ word-piece,
and the efficacy of different attacks as add $>$ key $>$ drop $>$ swap.

Next, we scrutinize the effectiveness of defense methods
when faced against adversarially chosen attacks.
Clearly from table~\ref{tab:sent-results},
DA and Adv
are not effective in this case.
We observed that despite a low training error,
these models were not able to generalize to attacks
on newer words at test time.
ATD spell corrector is the most effective on keyboard attacks,
but performs poorly on other attack types,
particularly the add attack strategy.

The ScRNN model
with pass-through backoff offers better protection,
bringing back the adversarial accuracy within $5\%$ range
for the swap attack.
It is also effective under other attack classes,
and can mitigate the adversarial effect
in word-piece models by $21\%$,
character-only models by $19\%$,
and in word, and word+char models by over $4.5\%$ .
This suggests that the
direct training signal of word error correction
is more effective than the indirect signal of sentiment
classification available to DA and Adv for model
robustness.

We observe additional gains by using background models as a backoff alternative,
because of its lower word error rate (WER), especially,
under the swap and drop attacks.
However, these gains do not consistently translate in all other settings,
as lower WER is necessary but not sufficient.
Besides lower error rate, we find that
a solid defense should furnish the attacker the fewest options to attack,
i.e. it should have a low sensitivity.

As we shall see in section \S~\ref{subsub:sensitivity}, the backoff neutral variation has the lowest sensitivity
due to mapping \unkspace predictions to a fixed neutral word.
Thus, it results in the highest robustness on most of the attack types
for all four model classes.

\begin{table}[ht]
\small
\centering
\begin{tabular}{@{}lccc@{}}
\toprule
\multirow{2}{*}{\textbf{Model}} & \multirow{2}{*}{\textbf{No Attack}} & \multicolumn{2}{c}{\textbf{All attacks}} \\ \cmidrule(l){3-4}
                                &                                     & \textbf{1-char}     & \textbf{2-char}    \\ \midrule
BERT                            & 89.0                                & 60.0                & 31.0               \\
BERT + ATD                      & 89.9                                & 75.8                & 61.6               \\
BERT + Pass-through             & 89.0                                & \textbf{84.5}       & 81.5               \\
BERT + Neutral                  & 84.0                                & 82.5                & \textbf{82.5}      \\ \bottomrule
\end{tabular}
\caption{Accuracy of BERT, with and without defenses, on MRPC when attacked under the `all' attack setting.
}
\label{tab:bert}
\end{table}

Table~\ref{tab:bert} shows the accuracy of BERT on $200$ examples
from the dev set of the MRPC paraphrase detection task
under various attack and defense settings.
We re-trained the ScRNN model variants on the MRPC training set for these experiments.
Again, we find that simple $1$-$2$ character attacks
can bring down the accuracy of BERT significantly ($89\%$ to $31\%$).
Word recognition models can provide an effective defense,
with both our pass-through and neutral variants recovering most of the accuracy.
While the neutral backoff model is effective on $2$-char attacks,
it hurts performance in the \emph{no attack} setting,
since it incorrectly modifies certain correctly spelled entity names.
Since the two variants are already effective,
we did not train a background model for this task.

\begin{table}
\centering
\small
\begin{tabular}{lccccc}
     \multicolumn{6}{c}{\textbf{Sensitivity Analysis}}\\
     \toprule
    \bf{Backoff}  & \bf{Swap} & \bf{Drop} & \bf{Add} & \bf{Key} & \bf{All} \\
    \midrule
    \multicolumn{6}{c}{Closed Vocabulary Models (word-only)}\\
    \midrule
    Pass-Through & 17.6 & 19.7 & 0.8 & 7.3 & 11.3 \\
    Background & 19.5 & 22.3 & 1.1 & 9.5 & 13.1 \\
    Neutral & 17.5 & 19.7 & 0.8 & 7.2 & 11.3 \\
    \midrule
    \multicolumn{6}{c}{Open Vocab. Models (char/word+char/word-piece)}\\
    \midrule
    Pass-Through & 39.6 & 35.3 & 19.2 & 26.9 & 30.3 \\
    Background & 20.7 & 25.1 & 1.3 & 11.6 & 14.7\\
    Neutral &  17.5 & 19.7 & 0.8 & 7.2 & 11.3\\
\bottomrule
\end{tabular}
\caption{
Sensitivity values for word recognizers.
Neutral backoff shows lowest sensitivity.
}
\label{tab:sensitivity}
\end{table}

\subsection{Understanding Model Sensitivity}
\label{subsub:sensitivity}
\paragraph{Experimental setup}
To study model sensitivity,
for each sentence,
we perturb one randomly-chosen word and replace it
with all possible perturbations under a given attack type.
The resulting set of perturbed sentences
is then fed to the word recognizer (whose sensitivity is to be estimated).
As described in equation~\ref{eq:sensitivity},
we count the number of unique predictions from the output sentences.
Two corrections are considered unique
if they are mapped differently by the downstream classifier.

\paragraph{Results}
The neutral backoff variant has the lowest sensitivity (Table~\ref{tab:sensitivity}).
This is expected, as it returns a fixed neutral word
whenever the ScRNN predicts an \unk,
therefore reducing the number of unique outputs it predicts.
Open vocabulary (i.e. char-only, word+char, word-piece) downstream classifiers
consider every unique combination of characters differently,
whereas word-only classifiers internally treat all out of vocabulary (OOV) words alike.
Hence, for char-only, word+char, and word-piece models,
the pass-through version is more sensitive than the background variant,
as it passes words as is (and each combination is considered uniquely).
However, for word-only models, pass-through
is less sensitive as all the OOV character combinations are rendered identical.

\begin{figure}
    \centering
   \includegraphics[width=0.49\linewidth, trim={0 1.5in 0.4in 1in}, clip]{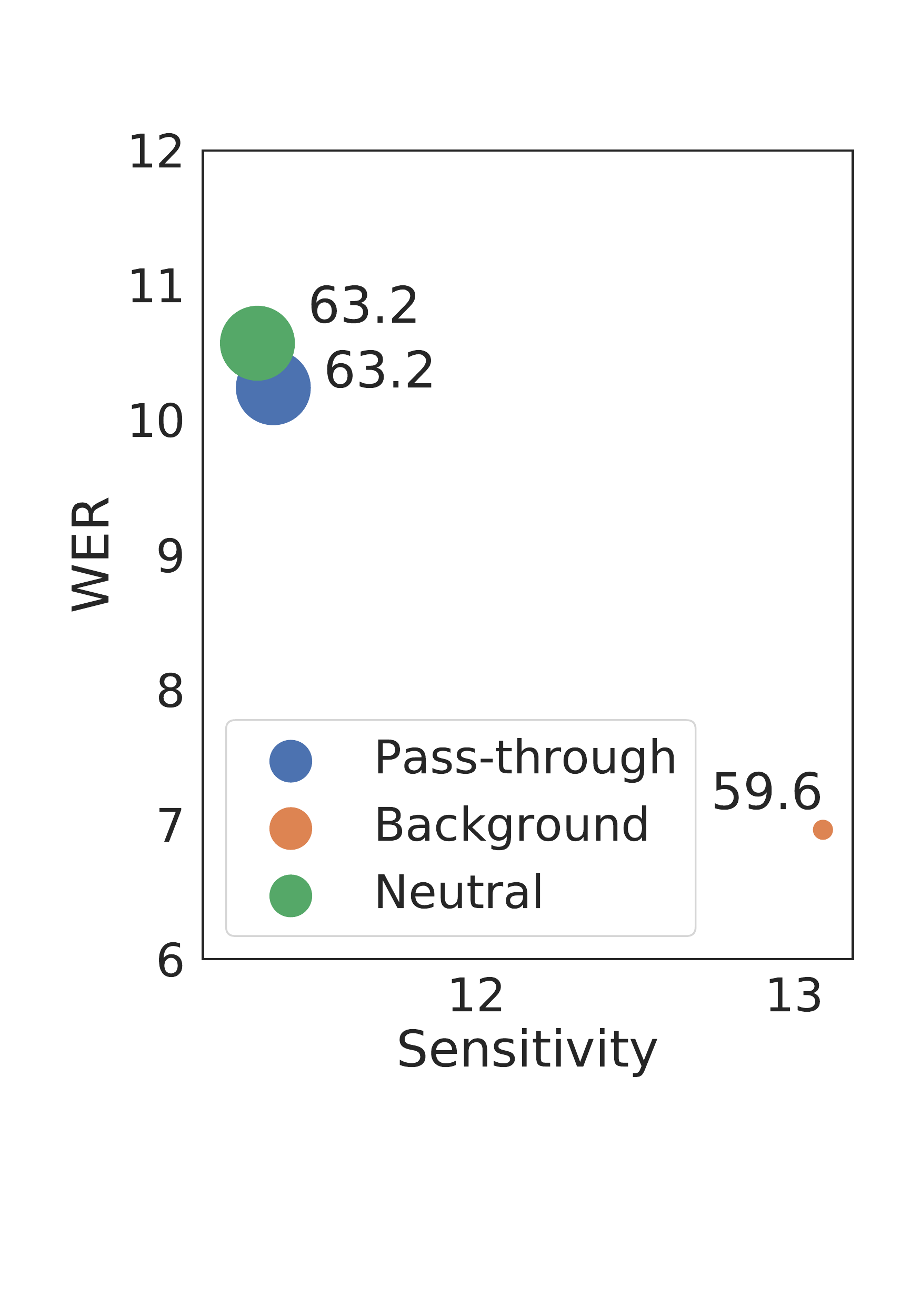}
    \includegraphics[width=0.48\linewidth, trim={0 1.5in 0.4in 0.8in}, clip]{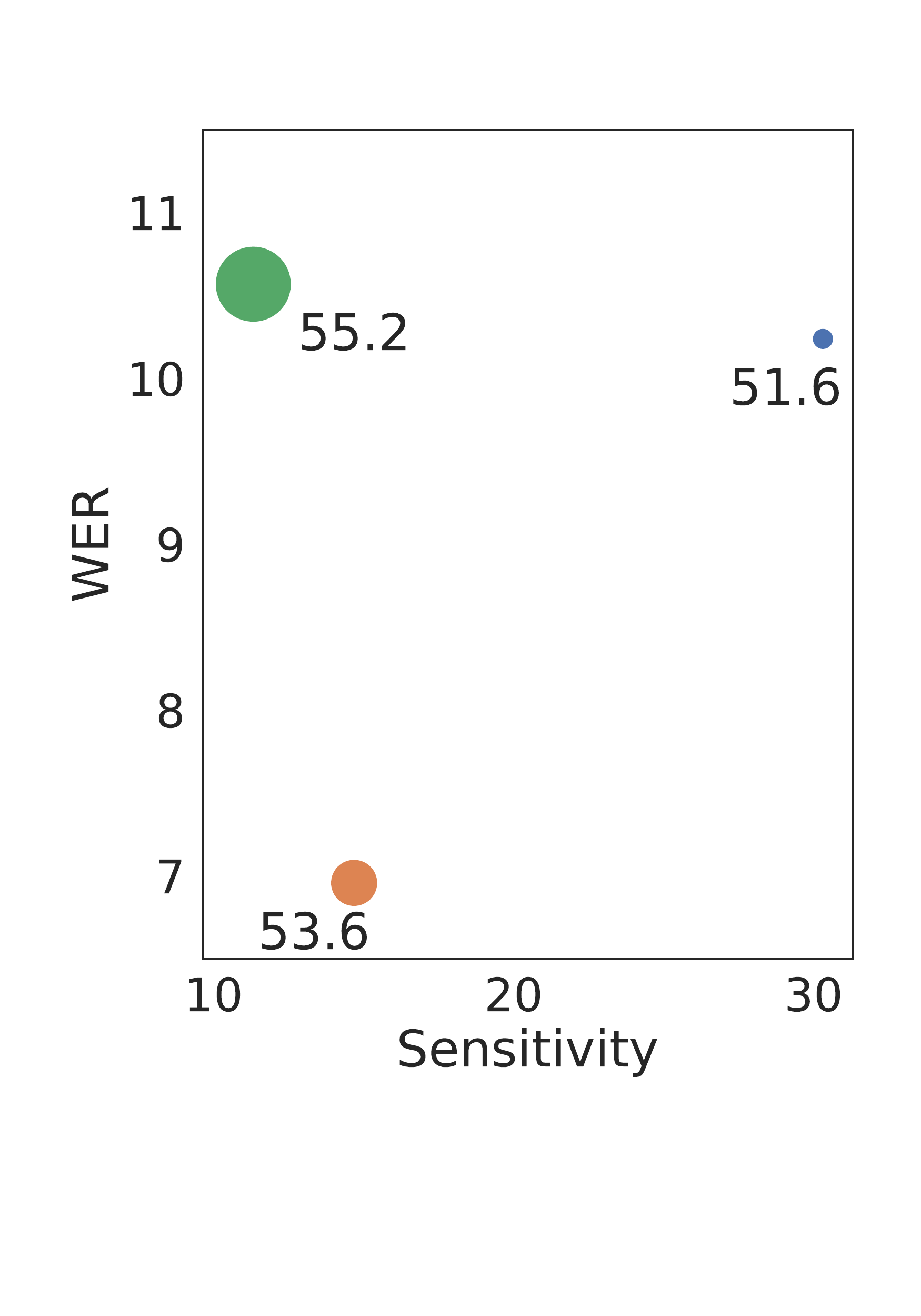}
\caption{
Effect of sensitivity and word error rate on robustness (depicted by the bubble sizes) in word-only models (left) and char-only models (right).
}
\vspace{-0.15in}
\label{fig:sensitivity}
\end{figure}

Ideally, a preferred defense is one with low sensitivity and word error rate.
In practice, however, we see that a low error rate often comes at the cost of sensitivity.
We see this trade-off
in Figure~\ref{fig:sensitivity},
where we plot WER and sensitivity on the two axes,
and depict the robustness when using different backoff variants.
Generally, sensitivity is the more dominant factor
out of the two, as the error rates of the considered variants are reasonably low.

\paragraph{Human Intelligibility}
We verify
if the sentiment (of the reviews) is preserved with char-level attacks.
In a human study with $50$ attacked (and subsequently misclassified), and 50 unchanged reviews, it was noted that $48$ and $49$, respectively, preserved the sentiment.

\section{Conclusion}
\label{sec:conclusion}
As character and word-piece inputs become commonplace in modern NLP pipelines,
it is worth highlighting the vulnerability they add.
We show that minimally-doctored attacks
can bring down accuracy of classifiers to random guessing.
We recommend word recognition as a safeguard against this
and build upon RNN-based semi-character word recognizers.
We discover that when used as a defense mechanism,
the most accurate word recognition models
are not always the most robust against adversarial attacks.
Additionally, we highlight the need
to control the sensitivity of these models to achieve high robustness.

\section{Acknowledgements}
\label{sec:ack}
The authors are grateful to Graham Neubig,
Eduard Hovy, Paul Michel, Mansi Gupta, and Antonios Anastasopoulos
for suggestions and feedback.
We thank Salesforce Research for their generous support
for our work on robust deep learning under distribution shift.

\bibliographystyle{acl_natbib}
\bibliography{acl2019.bib}

\end{document}